%% file: main.tex
\documentclass[letterpaper, 10 pt, conference]{ieeeconf}
\IEEEoverridecommandlockouts  
\usepackage{hyperref}
\usepackage{cite}
\usepackage{authblk}
\usepackage{amsmath,amssymb,amsfonts}
\usepackage[titlenumbered, ruled, linesnumbered, vlined]{algorithm2e}
\usepackage{graphicx}
\usepackage{subcaption}
\usepackage{array} 
\usepackage{cellspace}  
\usepackage{makecell}  
\usepackage{multirow} 
\usepackage{multicol}
\usepackage{booktabs}
\usepackage{siunitx}
\usepackage{caption,subcaption}
\usepackage{textcomp}
\usepackage{xcolor, colortbl}
\usepackage{todonotes,soul}
\usepackage{multirow}
\def\BibTeX{{\rm B\kern-.05em{\sc i\kern-.025em b}\kern-.08em
    T\kern-.1667em\lower.7ex\hbox{E}\kern-.125emX}}
\usepackage{amsmath,amssymb}
\let\oldnl\nl
\newcommand{\nonl}{\renewcommand{\nl}{\let\nl\oldnl}}
\def\fixme#1{\typeout{FIXED in page \thepage : {#1}}
\bgroup \color{red}{[FIXME: {#1}]} \egroup}
\usepackage{changes}

\begin{document}
\title{TinyLidarNet: 2D LiDAR-based End-to-End Deep Learning Model for F1TENTH Autonomous Racing}
\author{Mohammed Misbah Zarrar}
\author{Qitao Weng}
\author{Bakhbyergyen Yerjan}
\author{Ahmet Soyyigit}
\author{Heechul Yun}
\affil{University of Kansas, Lawrence, KS\\ \{zarrar\_1607, wengqt, yerjanb, ahmet.soyyigit,  heechul.yun\}@ku.edu}

\maketitle

\begin{abstract}
Prior research has demonstrated the effectiveness of end-to-end deep learning for robotic navigation, where the control signals are directly derived from raw sensory data. However, the majority of existing end-to-end navigation solutions are predominantly camera-based. In this paper, we introduce TinyLidarNet, a lightweight 2D LiDAR-based end-to-end deep learning model for autonomous racing. 
An F1TENTH vehicle using TinyLidarNet won 3rd place in the 12th F1TENTH Autonomous Grand Prix competition, demonstrating its competitive performance. We systematically analyze its performance on untrained tracks and computing requirements for real-time processing. We find that TinyLidarNet's 1D Convolutional Neural Network (CNN) based architecture significantly outperforms widely used Multi-Layer Perceptron (MLP) based architecture. 
In addition, we show that it can be processed in real-time on low-end micro-controller units (MCUs). 

\end{abstract}



\input{sections/intro}
\input{sections/background}
\input{sections/f1tenth}

\input{sections/TinyLidarNet}
\input{sections/evaluation}
\input{sections/conclusion}

\input{sections/acknowledgement}
\bibliography{references}
\end{document}

%% file: sections/intro.tex
\section{Introduction}\label{sec-intro}




In F1TENTH autonomous racing~\cite{upennf1tenth,o2020f1tenth}, developing an intelligent, yet computationally efficient control algorithm is necessary and challenging due to constraints in size, weight, and power. These systems must swiftly process sensor input data to make real-time decisions and allow fast collision-free navigation of the ego-vehicle in various racing tracks.
Traditional approaches, which involve a complex pipeline of perception, planning, and control algorithms, are challenging to apply in fast-paced autonomous racing due to high computational costs and sensitivity to perception and modeling errors, which can propagate and accumulate through the pipeline
End-to-end (E2E) deep learning approaches~\cite{Levine2016,Bojarski2016,liu2021efficient,chen2023endtoend} present a promising alternative as they can simplify the control pipelines, improve computational efficiency, and achieve high performance~\cite{chen2023endtoend}.  

In end-to-end approaches, a neural network replaces all or a subset of perception, planning, and control algorithms in the traditional control pipeline~\cite{Levine2016,betz2022autonomous}. 
Many prior work has explored end-to-end approaches for autonomous driving~\cite{Bojarski2016,pilotnet20,pan2019agile}. 
However, the majority of these prior works are vision-based approaches, which require a large amount of training data to achieve good performance and consistency and are susceptible to environmental factors such as lighting conditions~\cite{pilotnet20,Bojarski2016}.
Several studies have investigated end-to-end approaches based on 2D LiDAR (Light Detection and Ranging) in the context of F1TENTH racing specifically~\cite{chen2023endtoend,codevilla2018endtoend,wadekar2021endtoend}. Compared to cameras, 2D LiDARs generate fewer data, making the training of 2D LiDAR models easier. 

Most prior 2D LiDAR-based end-to-end approaches for F1TENTH racing are based on multi-layer perceptron (MLP) models 
that take a 2D LiDAR scan (e.g., a 1081 dimensional vector over a 270-degree field of view) as input and predict control commands (e.g., throttle and steering) as output~\cite{sun2023benchmark}.
However, these MLP-based models do not perform well at high speeds and do not generalize well across different environments~\cite{sun2023benchmark}. 

In this work, we aim to address the following research questions:
(1) Can we develop a 2D LiDAR-based end-to-end deep learning model that demonstrates competitive performance in actual F1TENTH autonomous racing competitions?
(2) What level of computational power is needed to execute such a deep learning model and achieve competitive performance in racing? 
(3) Can we achieve good performance on unseen tracks, both in the real world and in simulation, without additional data collection and retraining?
In other words, how well does our end-to-end model generalize, especially compared to prior MLP-based models? 

To this end, 
first, we present TinyLidarNet, an end-to-end deep convolutional neural network (CNN) that directly takes a raw 2D LiDAR scan as input and predicts throttle and steering control commands. 
TinyLidarNet is inspired by the PilotNet architecture~\cite{Bojarski2016}, a vision-based end-to-end CNN model used in NVIDIA's Dave 2 project to drive a real car in various real road conditions. Instead of using 2D convolutions, as in the original PilotNet, TinyLidarNet uses 1D convolutions to capture the spatial features of the incoming 2D LiDAR scans. TinyLidarNet shows competitive performance in an actual F1TENTH competition~\footnote{3rd place winner at \href{https://cps2023-race.f1tenth.org}{12th F1TENTH autonomous grand-prix competition}} with a relatively small amount of training data
collected on the competition track (Figure~\ref{fig:12th_f1tenth_track}), following the standard behavior cloning approach~\cite{Levine2016}. 
Second, we find that TinyLidarNet is computationally efficient
and can perform an inference in less than 1 millisecond on the NVIDIA Jetson NX platform. We further find that it is even possible to execute the network on a tiny MCU in real-time. Specifically, we port the TinyLidarNet to an ESP32-S3 and Raspberry Pi Pico MCU and can achieve sub-50ms latencies ($>$20Hz) after applying a standard 8-bit integer quantization using TensorFlowLite-Micro~\cite{david2020tensorflow}.
Lastly, we show that the trained model generalizes well on unseen tracks, both in the real world as well as in simulation, even without any additional rounds of data collection and augmentation regimens such as DAgger~\cite{ross2011reduction}. This is in stark contrast with prior MLP-based 2D LiDAR models~\cite{chen2023endtoend,codevilla2018endtoend,wadekar2021endtoend}, which struggle to work well at high speed and unseen tracks~\cite{sun2023benchmark}. This is because TinyLidarNet's 1D convolutional filters can better capture the spatial features of the 2D LiDAR scans.


In summary, we make the following contributions: 
\begin{itemize}
 \item We present TinyLidarNet, a 2D LiDAR-based end-to-end CNN architecture for F1TENTH autonomous racing, which shows competitive performance in real racing and generalizes well on unseen tracks.  We release the code and training data as open source~\footnote{
 \url{https://github.com/CSL-KU/TinyLidarNet}
 }.  
 \item We systematically explore the latency, accuracy, and performance trade-offs of TinyLidarNet on contemporary embedded computing platforms. We show the feasibility of using a tiny MCU to run TinyLidarNet in real-time. 
 \item We provide extensive evaluation results of TinyLidarNet's performance and generalizability, compared to the state-of-the-art 2D LiDAR-based end-to-end MLP models.
 We show that TinyLidarNet's CNN architecture is superior to conventional MLP architectures for processing 2D planner LiDAR input. 
 
\end{itemize}

%% file: sections/background.tex
\section{Background and Related Work}\label{sec:background}

\subsection{F1TENTH Autonomous Racing Competition}

\begin{figure}[htp]
  \centering
  \includegraphics[width=\linewidth]{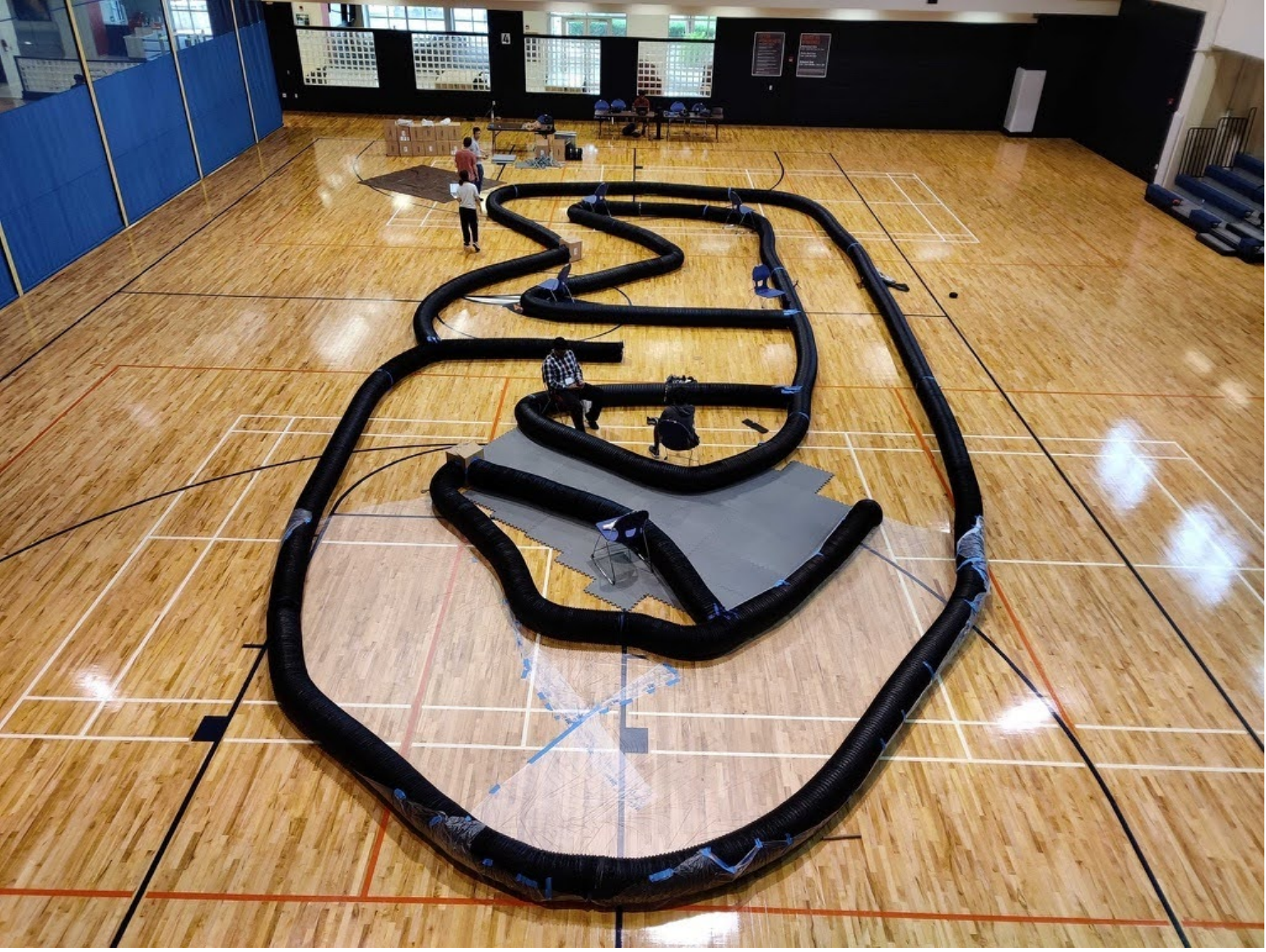}
  \caption{12th F1TENTH Grand Prix Competition Track}
  \label{fig:12th_f1tenth_track}
\end{figure}

The F1TENTH Autonomous Racing competition~\cite{upennf1tenth} is a competitive autonomous racing event to develop and test algorithms for autonomous racing with 1/10th scale race cars. 
It challenges the scaled race cars to autonomously navigate and complete races in a given race track as fast as possible without collision. 
These race cars come equipped with an array of sensors, but the 2D planner LiDAR is the most commonly used primary sensor for perception. 
The competition provides a realistic and dynamic testing ground for researchers and students in the field of autonomous vehicles, robotics, and artificial intelligence. 
Figure~\ref{fig:12th_f1tenth_track} shows the race track used in the 12th F1TENTH Grand Prix competition held in IEEE/ACM CPS-IoT Week 2023~\footnote{\url{https://cps2023-race.f1tenth.org}}. 

\subsection{End-to-end Deep Learning for Autonomous Driving}

Rather than relying on conventional modular systems, end-to-end approaches utilize deep neural networks, transforming sensor input data into control outputs~\cite{betz2022autonomous}.
The concept was initially introduced in the 1980s \cite{Pomerleau1989} with a compact 3-layer fully connected neural network tailored for autonomous cars. Subsequent to that, the DARPA Autonomous Vehicle (DAVE) project emerged in the early 2000s \cite{LeCun:04} and numerous other projects followed the suit~\cite{Bojarski2016,bechtel2018picar,bechtel2022deeppicarmicro,weiss2020deepracing}. An example is NVIDIA's DAVE-2 project, featuring a 9-layer CNN for end-to-end control, demonstrating its ability to drive a real car in various road conditions \cite{Bojarski2016}. 
Most prior works are vision-based, taking raw image pixels as input and directly generating control commands as output. 
In \cite{liu2021efficient}, on the other hand, a 3D LiDAR is used as the primary input sensor, which produces a 3D point cloud as input to the end-to-end model, in conjunction with GPS and coarse-grain map information. Their end-to-end model directly processes the 3D point cloud via 3D sparse convolutions and was shown to be able to produce robust steering controls in a real autonomous car.

Several prior works have investigated 2D planner LiDAR-based end-to-end approaches in the context of F1TENTH racing~\cite{chen2023endtoend,codevilla2018endtoend,wadekar2021endtoend,sun2023benchmark}. 
Compared to cameras or 3D LiDARs, the 2D planner LiDAR in F1TENTH racing cars (see Section~\ref{sec:f1tenth}) generates significantly fewer amount data, mainly the distances from the ego vehicle to the surrounding objects or walls on the 2D plane, which makes it easier to process. These prior works targeting F1TENTH racing are based on MLP models to process 2D LiDAR scans. 
However, in the F1TENTH community, end-to-end approaches have not been popular in actual competitions due to poor performance. For example, in the 12th F1TENTH Grand Prix, we were the only team that adopted an end-to-end approach. 
In~\cite{bosello2022train}, it is shown that a CNN model can be trained more efficiently than an MLP model in a deep reinforcement learning framework, but it did not show high-speed racing capability in the real world. 
In a recent survey by the creators of the F1TENTH racing~\cite{betz2022autonomous}, it is noted that end-to-end approaches exhibit low achievable speed, non-ideal driving characteristics (e.g., wobbly behavior), and low generalizability on unseen tracks, among other limitations. In this work, we challenge these preconceived weaknesses of end-to-end approaches in the context of F1TENTH Racing. 



%% file: sections/f1tenth.tex

\section{F1TENTH Platform and 2D LiDAR} 
\label{sec:f1tenth}

\begin{figure}[htp]
    \centering
    {\includegraphics[width=0.9\linewidth]{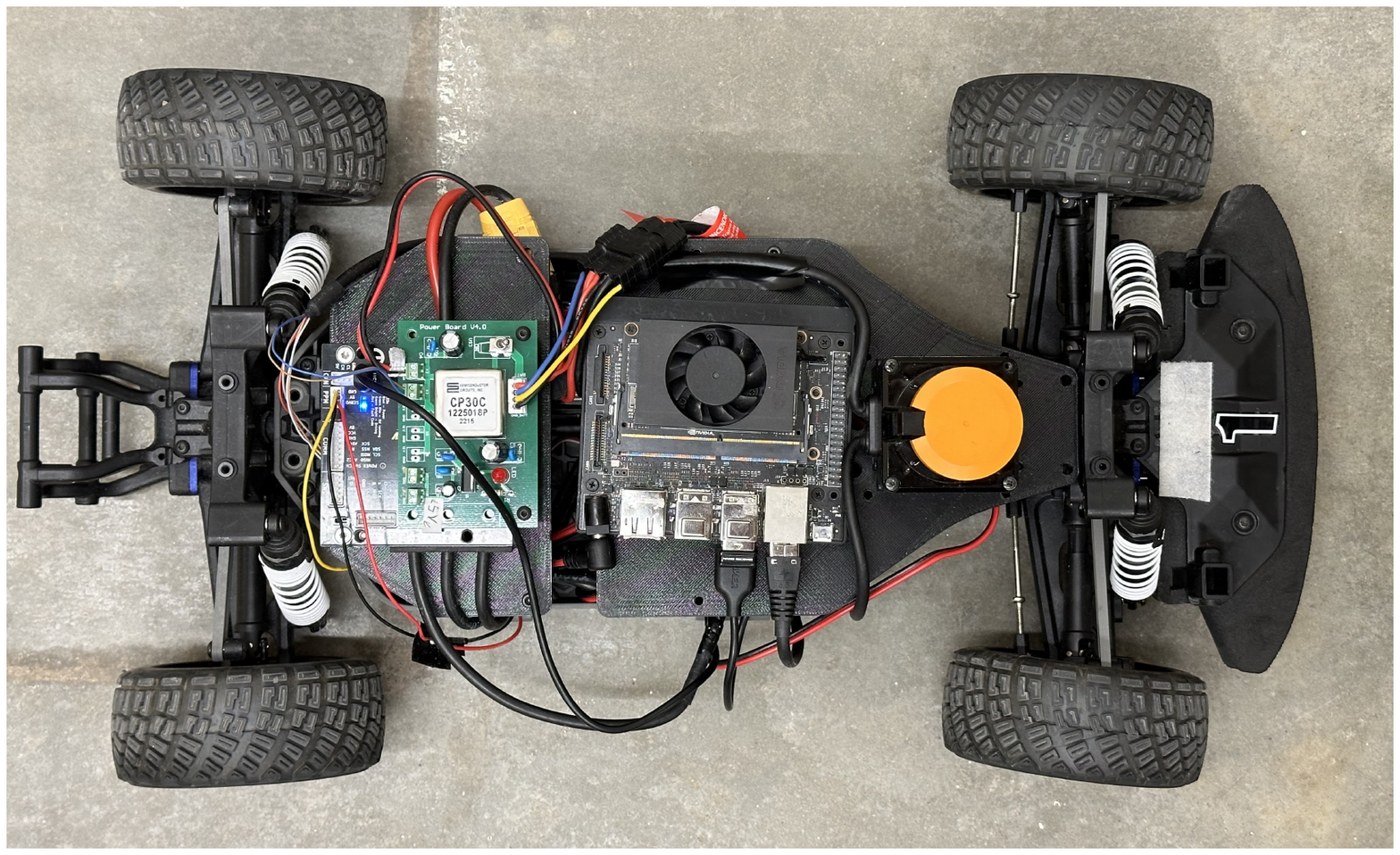}}
    \caption{F1TENTH platform with a Hokuyo UST-10LX 2D LiDAR and a NVIDIA Jetson Xavier NX on-board computer.}
    \label{fig:f1tenth_harware_setup}
\end{figure}


Figure \ref{fig:f1tenth_harware_setup} shows the F1TENTH car utilized in all of our real-world experiments.
The chassis of the car is based on a Traxxas Rally 1/10-scale radio-controlled car with an Ackermann steering mechanism. 
Its primary sensor is a Hokuyo UST-10LX 2D Planner LiDAR, which has a \ang{270} field of view with a range of up to 10 meters. This LiDAR device provides an angular resolution of \ang{0.25} and operates at a scan frequency of 40Hz, generating 1081 data points represented as a 1D distance array, which serves as the input for our end-to-end model. 
Since the scan frequency is 40 Hz, naturally the deadline for prediction of speed and throttle becomes 0.025 seconds or 25 milliseconds. For control, an open-source electronic speed control (ESC) board controls the RPM of a brushless motor and a steering servo. A power board distributes power from a lithium polymer (LiPO) battery to the sensors, motors, and the on-board computer. For on-board computing, an NVIDIA Jetson Xavier NX embedded computer is used to run all software, which equips 8 64-bit ARM CPU cores and a 384-core Volta GPU. 
On the software side, we use the ROS Noetic Ninjemys framework on Ubuntu 20.04.6 operating system. 



%% file: sections/TinyLidarNet.tex
\section{TinyLidarNet} \label{sec:TinyLidarNet}

In this section, we introduce the TinyLidarNet architecture and our data collection and training process. 

\subsection{Architecture} \label{sec:TinyLidarNet_Architect}

\begin{figure}[htp]
\centering
\includegraphics[width=0.75\linewidth]{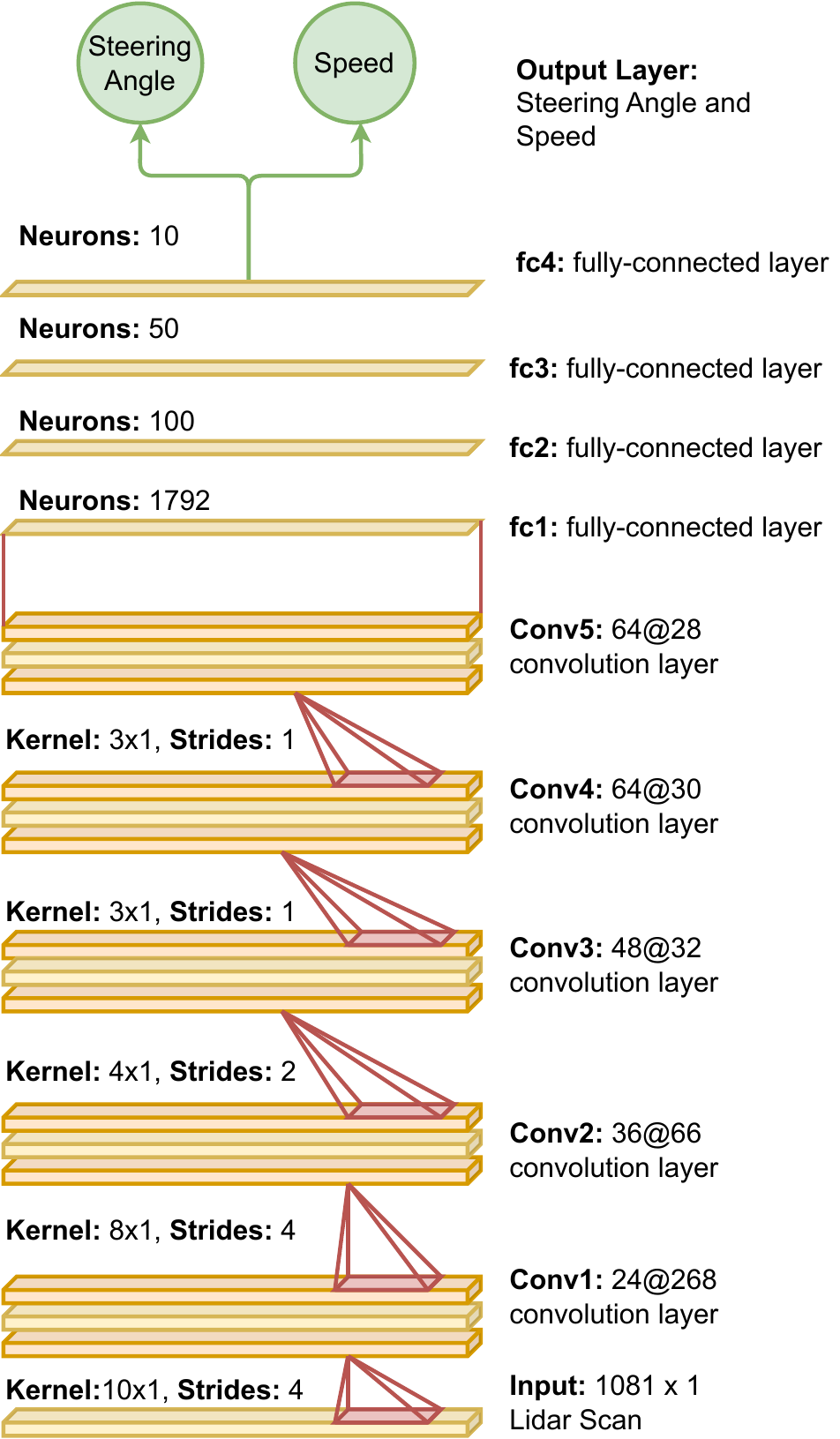}
\caption{TinyLidarNet architecture:  9 layers (5 convolutional, 4 fully-connected) with 220,686 parameters.}
\label{fig:TinyLidarNet_Architecture}
\end{figure}

Figure \ref{fig:TinyLidarNet_Architecture} shows the architecture of TinyLidarNet. The architecture is comprised of 9 layers (5 convolutional and 4 fully connected) with a total of 220K parameters and takes about 1.5 million multiply-accumulate operations (MACs) to execute. 
The architecture is inspired by the PilotNet architecture~\cite{Bojarski2016}, which is a vision-based end-to-end model for NVIDIA's Dave2 self-driving car. The main differences are that it takes a 2D LiDAR scan, instead of an RGB image, as input and uses 1D CNN layers instead of 2D ones. 
Quantitatively, compared to PilotNet~\cite{Bojarski2016}, TinyLidarNet has a similar number of parameters (250K vs. 220K) but requires significantly lower MACs (27 million vs. 1.5 million) because the input dimensionality is much lower (200x66 RGB pixels vs. 1081 LiDAR scan samples). As such, it has a significantly lower computational demand, which makes it possible to use less powerful computing platforms, as we investigate in Section~\ref{sec:eval-inference_latency}.

\subsection{Data Collection, Pre-processing and Training} \label{sec:TinyLidarNet_Training}


The data collection is carried out by driving the vehicle manually using a joystick.
The collected data  
includes servo angle and 
speed data from the ESC board of the car
and the 1D range array (1081 values) from the Hokuyo 2D LiDAR.

We collected a total of 12,329 samples, approximately 5 minutes of driving on the training track, shown in Figure~\ref{fig:12th_f1tenth_track}. 
Note that during the data collection, the car drove with stationary obstacles (opponent vehicles) present. 
We assigned 85\% (10,478 samples) of the collected samples to training and the remaining portion to testing and validation. We used a batch size of 64 and trained the network for 20 epochs with a learning rate of 5e-5 and Huber loss~\cite{Huber1964RobustEO}.

The TinyLidarNet training process is performed locally on the NVIDIA Jetson Xavier NX platform and takes approximately 4 minutes to complete. 





During the data collection process, we found that the quality of the LiDAR scan samples deteriorated significantly when the car navigated dark track sections. LiDAR sensors tend to perform less effectively on black surfaces, resulting in noisy point clouds~\cite{pericles_264045673}.
To reduce noise, we employ median and interpolation filters. 
These filters play a crucial role in stabilizing point clouds. 

%% file: sections/evaluation.tex
\section{Evaluation}\label{sec:evaluation}
In this section, we evaluate the performance of TinyLidarNet. 

\subsection{Insights from an F1TENTH Competition}
\label{sec:eval-insight_from_competition}


Using TinyLidarNet, we participated in the 12th F1TENTH Autonomous Grand Prix competition and won the 3rd place award out of 13 participating teams. Although TinyLidarNet did not secure the top position, it exhibited competitive performance and demonstrated several advantages over traditional approaches that relied on perception, planning, and control pipelines.

First, during the competition, the track depicted in Figure \ref{fig:12th_f1tenth_track} underwent frequent alterations due to collisions. Each time a racecar collided with the track, the track was manually adjusted, leading to slight changes in its configuration.  This posed a considerable challenge for participants who relied on an offline map for path planning.
In contrast, TinyLidarNet does not rely on an offline map and was largely immune to the alterations in the track configuration. 

Second, a particularly interesting capability of TinyLidarNet is its ability to overtake other cars during head-to-head races in the tournament. This is interesting because 
the training dataset comprises examples of the car driving on the competition track by a human driver with static obstacles only, as detailed in Section~\ref{sec:TinyLidarNet_Training}. 
Note that executing overtaking maneuvers in high-speed racing is a challenging task for classical control approaches~\cite{bak2022stress,zhang2023f1tenth}, and most participants in the competition did not attempt to execute overtaking maneuvers during the race. A team that implemented an overtaking algorithm did so only in the long straight section of the track (Figure \ref{fig:12th_f1tenth_track}). 
On the other hand, TinyLidarNet was able to overtake moving opponents in any part of the track throughout the competition, likely because moving race cars were recognized as static obstacles or walls.

Overall, the competitive performance of TinyLidarNet in the competition motivated us to systematically analyze its performance and characteristics.


\subsection{Performance on Simulated Tracks } \label{sec:eval-generalizability}

In this subsection, we evaluate TinyLidarNet's performance on unseen virtual tracks.

\begin{figure}[htp]  
  \begin{subfigure}{0.45\linewidth}
    \includegraphics[width=\linewidth]{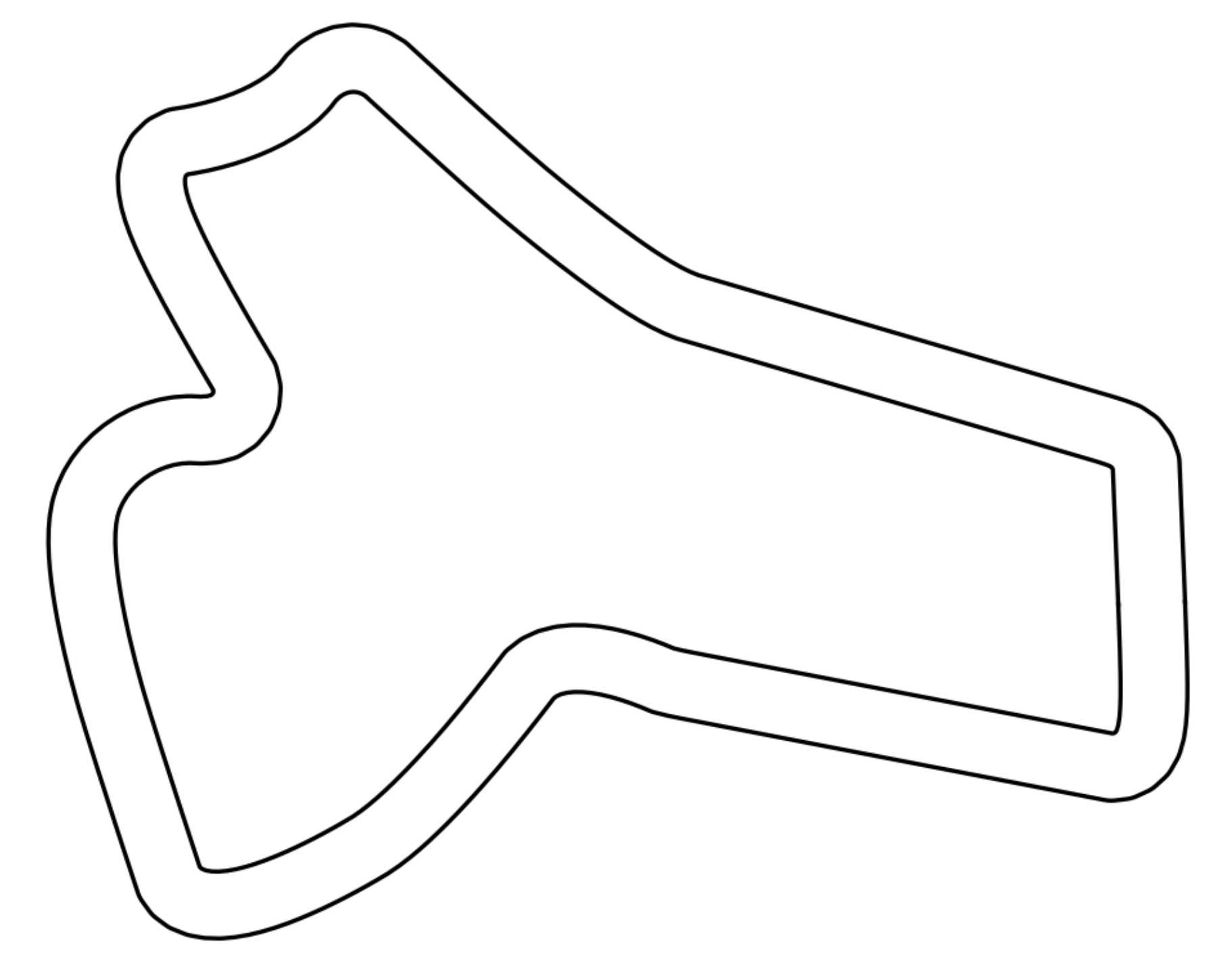}
    \caption{F1TENTH Gym (GYM)}
    \label{fig:f1tenth_gym}
  \end{subfigure}
  \begin{subfigure}{0.45\linewidth}
    \includegraphics[width=\linewidth]{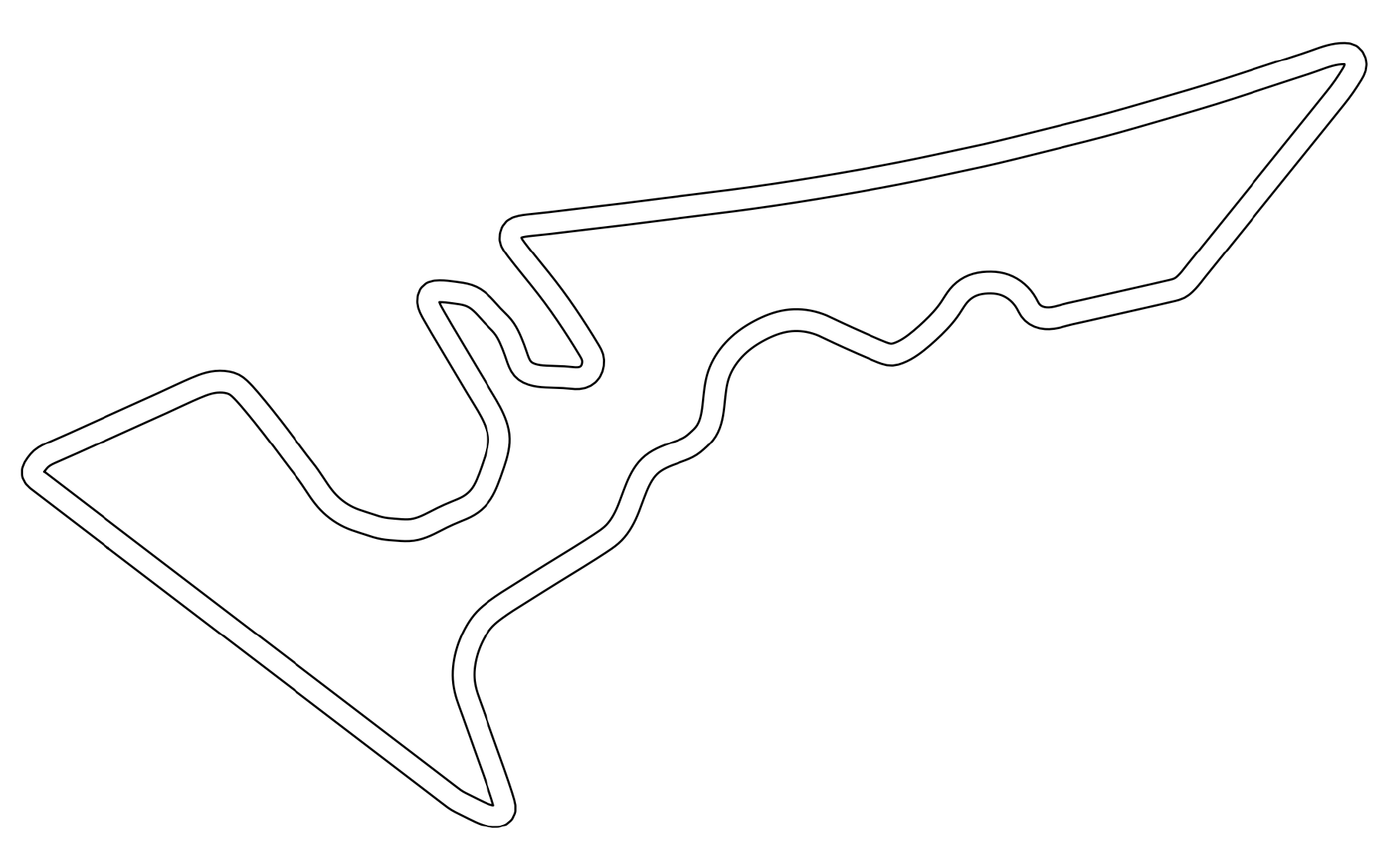}
    \caption{Austin (AUS)}
    \label{fig:Austin}
  \end{subfigure}
  \begin{subfigure}{0.45\linewidth}
    \includegraphics[width=\linewidth]{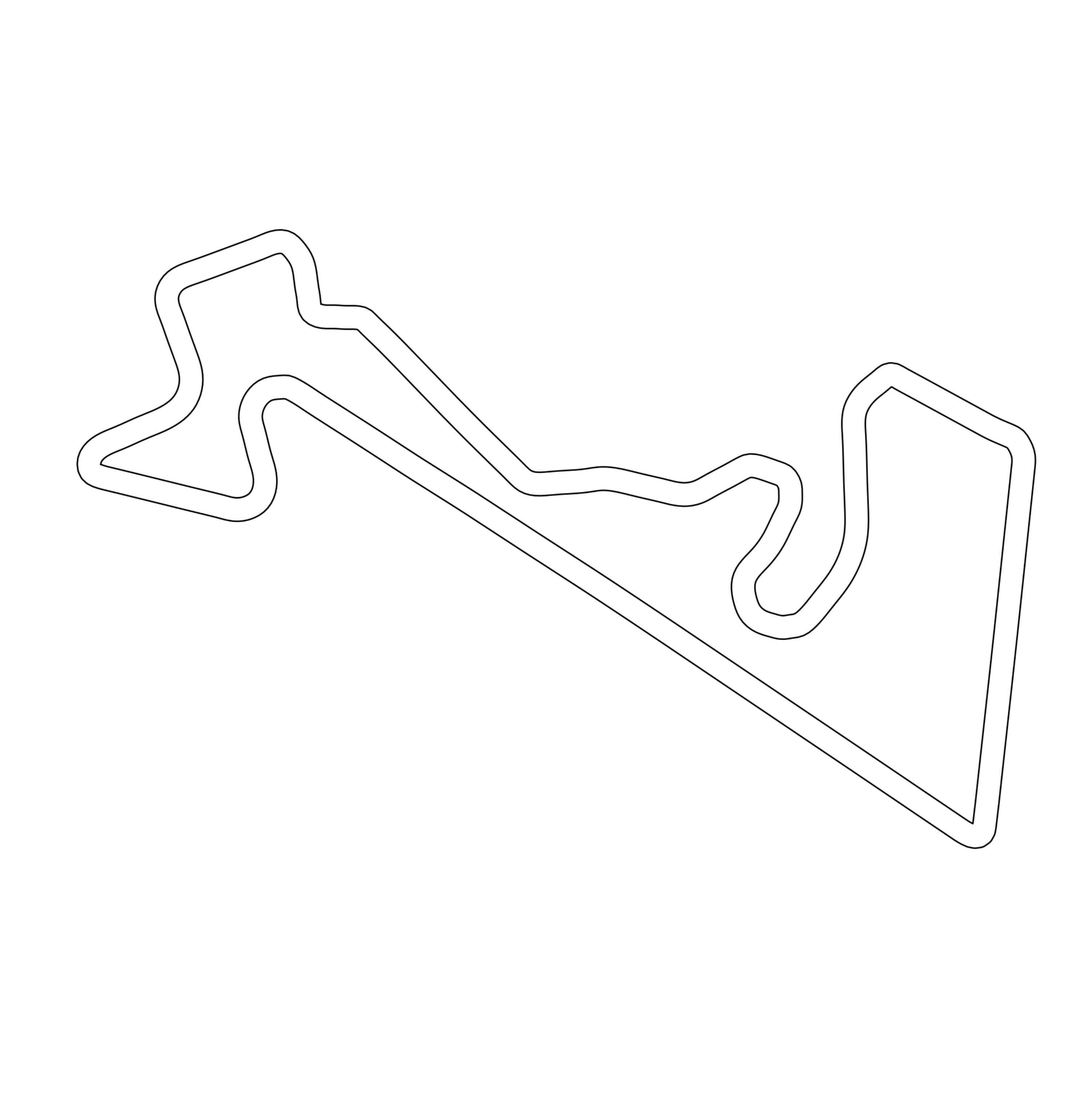}
    \caption{Moscow (MOS)}
    \label{fig:S}
  \end{subfigure}
  \begin{subfigure}{0.45\linewidth}
    \includegraphics[width=\linewidth]{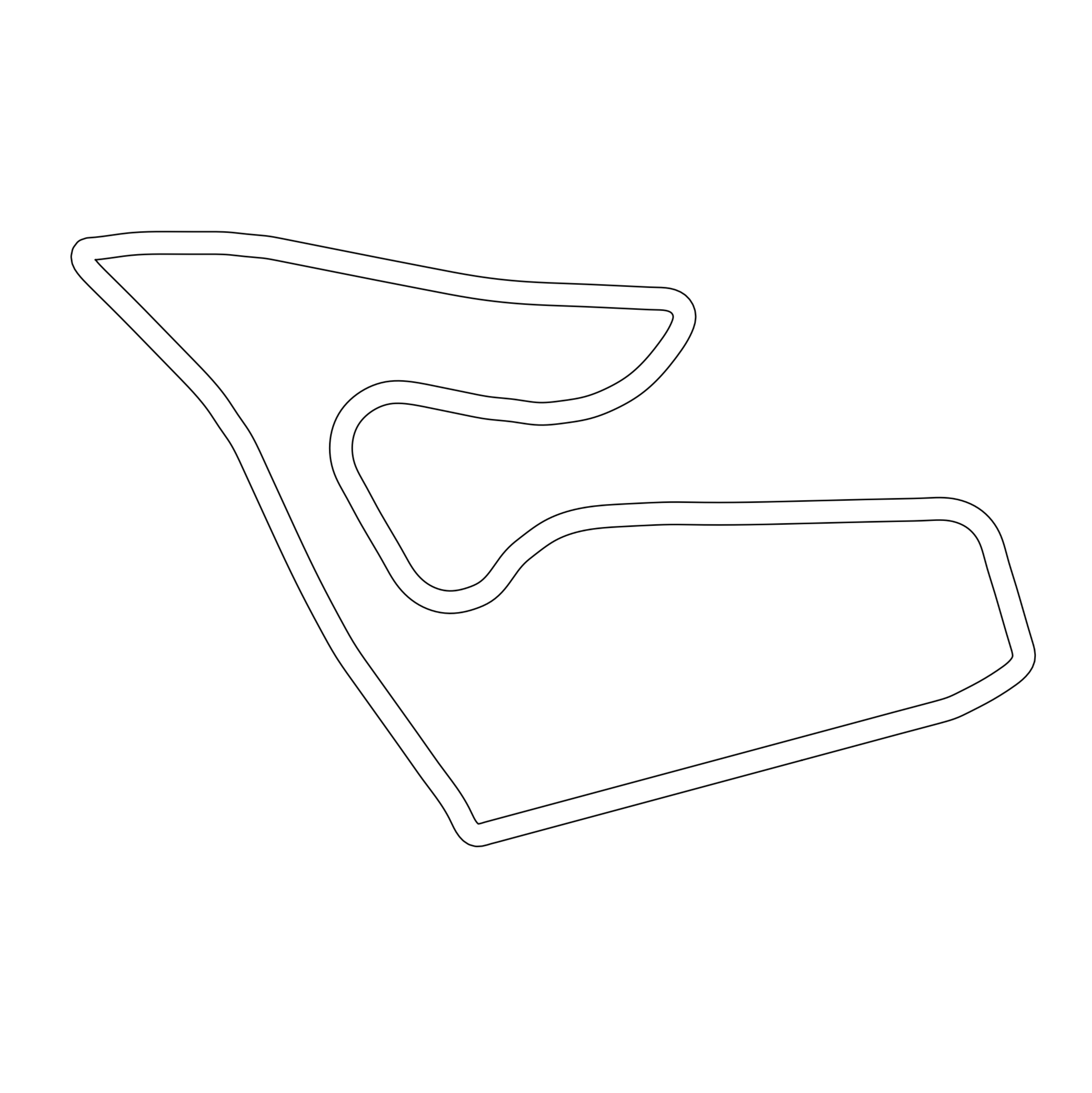}
    \caption{Spielberg (SPL)}
    \label{fig:Spielberg}
  \end{subfigure}
  \caption{Simulation Tracks from F1TENTH gym and F1TENTH racetrack repository\cite{okelly2020f1tenth,Betz2022_RacingSurvey}}
  \label{fig:Simulated Tracks}
\end{figure}

\begin{table}[htp]
    \centering
    \begin{tabular}{lccc}
    \toprule
        \textbf{Model} &\textbf{Input dim.}  & \textbf{Parameters} & \textbf{MACs}  \\
    \midrule
        TinyLidarNet$^{L}$ & 1081 & \makecell{220,686} & \makecell{1,546,960} \\
        TinyLidarNet$^{M}$ & 541  &  111,886 & 687,680 
        \\
         TinyLidarNet$^{S}$ & 271 &  \makecell{54,286}   & \makecell{240,752} 
        \\
        \hline
        MLP256$^L$~\cite{sun2023benchmark} & 1081  &  \makecell{343,298} &  \makecell{342,784} \\
        MLP256$^M$ & 541  &  \makecell{205,058} &  \makecell{204,544} \\
        MLP256$^S$ & 271  &  \makecell{135,938} &  \makecell{135,424} \\

        \bottomrule
    \end{tabular}
    \caption{Comparison of end-to-end models. 
    }
    \label{tab:MLP_comparision}
\end{table}

Table~\ref{tab:MLP_comparision} shows the basic characteristics of the compared 2D Lidar-based end-to-end models. Note that \textit{TinyLidarNet$^L$} is the original version that we use in an actual competition~\ref{sec:eval-insight_from_competition}. It is also the largest one as it takes all 1081 range data of a LiDAR scan as input. \textit{TinyLidarNet$^M$} and \textit{TinyLidarNet$^S$}are smaller variants, which down-sample the LiDAR scan by taking one for every 2 or 4 values of the range data, respectively. Except for the input dimension, they have identical 9-layer architecture. 
On the other hand, \textit{MLP256}$^L$~\cite{sun2023benchmark} is a multi-layer perceptron with 2 hidden layers, with 256 neurons in each layer. MLP256$^M$ and MLP256$^S$ are its smaller variants with smaller input dimensions. 

Figure~\ref{fig:Simulated Tracks} shows the four tracks we use for evaluation. These tracks are widely used in F1TENTH literature~\cite{okelly2020f1tenth,Betz2022_RacingSurvey}, which we use without any modifications. 

The basic evaluation setup is as follows. First, we train all compared models using the same dataset, which was originally collected on the F1TENTH competition, as described in Section~\ref{sec:TinyLidarNet_Training}, and evaluate them on the four tracks using a simulator infrastructure in ~\cite{evans2024unifying}. Using the simulation infrastructure, each end-to-end model drives a simulated car for one complete lap on a track for 10 trials, each time starting from a random position on the track. For each successful completion among the 10 trials, the \textit{average lap time} was calculated. In addition, the \textit{average progress} rate measures the percentage of the track the model can progress, on average, during the 10 trials. 

To mimic real-world conditions, we adjusted the simulation so that the control frequency is set to 40 Hz. 
We also add random Gaussian noise in the range of 0 to 0.5 to the LiDAR scans and perform a range clipping at 10 m to mimic the behavior of Hokuyo UST-10LX LiDAR in the F1TENTH platform, which has a range of 10 m. 

Note that a common practice of training an end-to-end model for a robot involves starting in simulation and eventually deploying it in the real world. However, bridging the Simulation to Reality gap (Sim2Real gap) can pose significant challenges\cite{TRENTSIOS2022287,10242366}. 
On the other hand, our evaluation strategy can be described as Real2Sim as we train the models using the real-world track dataset and test them in a simulated environment. 

\begin{table*}[htp]
\centering
{%
\begin{tabular}{l|cccc|cccc}

 \toprule
& \multicolumn{4}{c|}{\textbf{Average Lap Time (s)}} & \multicolumn{4}{c}{\textbf{Average Progress (\%)}} \\ 

\textbf{Model}   &  \textbf{GYM} &  \textbf{AUS} & \textbf{MOS} & \textbf{SPL} &   \textbf{GYM} & \textbf{AUS} & \textbf{MOS} & \textbf{SPL}  \\
\midrule

TinyLidarNet$^{L}$ & 25.8  & 85.7 & 63.3 & 65.3 & 100 & 100 & 100 & 100\\ 

TinyLidarNet$^{M}$ & 25.3  & 80.0 & 59.5 & 61.5 & 100 & 100 & 100 & 100\\ 

TinyLidarNet$^{S}$ & 26.9 & 83.4  & 61.8 & 64.1 & 100 & 100 & 100 & 100\\ 


\hline


MLP256$^L$~\cite{sun2023benchmark} & N/A  & N/A & 58.8 & 58.3 & 31 & 16 & 42 & 61 \\

MLP256$^{M}$ & 28.4  & N/A & 64.3 & 65.7 & 100 & 17 & 58 & 78 \\

MLP256$^{S}$& 27.6 & N/A & N/A & 62.2 & 77 & 48 & 29 & 37  \\















\bottomrule
\end{tabular}%
}
\caption{Average lap time (of successful trials) and the average progress (of all trials). At each trial, the ego-vehicle's position in the track is randomly chosen. 
}
\label{tab:performance_metrics}
\end{table*}

Table~\ref{tab:performance_metrics} shows the evaluation results. Note first that all three TinyLidarNet$^{L,M,S}$ can complete all 10 laps without any collision, hence achieving 100\% average progress, in all four tested tracks. On the other hand, MLP256 models 
struggle to complete the laps. MLP256 models could complete the lab some of the times, but their success rates are not close to 100\% in most cases. The average lap times of MLP256 models are also somewhat slower than that of TinyLidarNet models in many cases.
Overall, the TinyLidarNet variants show better performance, especially in terms of average progress, than the MLP256 variants. 

\begin{figure}[htp]
    \centering
    \rotatebox{0}{\includegraphics[width=1\linewidth]{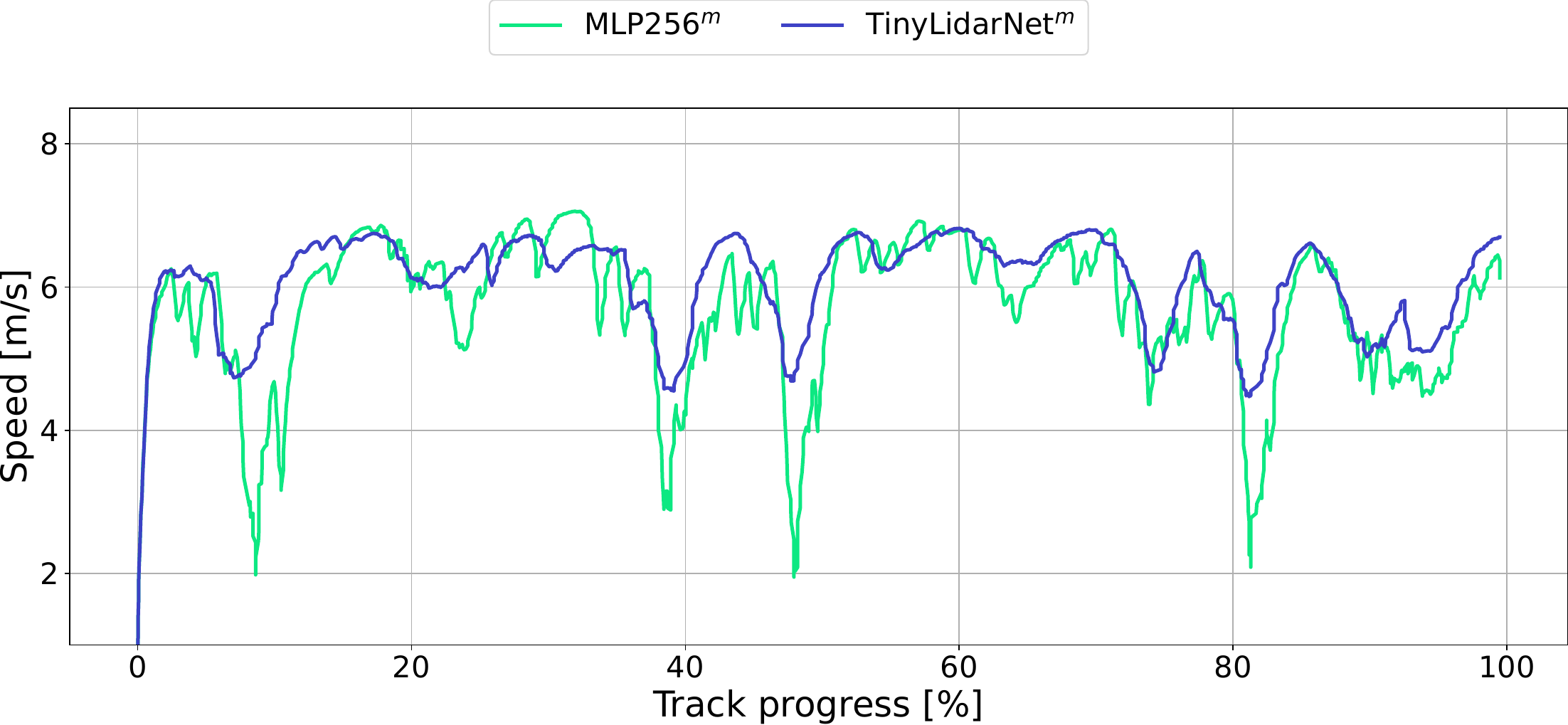}}
     \caption{Speed comparison of different models on GYM Track 
     }
    \label{fig:Compare_diff_tech}
\end{figure}


\begin{figure}[htp]
    \centering
    \rotatebox{0}{\includegraphics[width=0.82\linewidth]{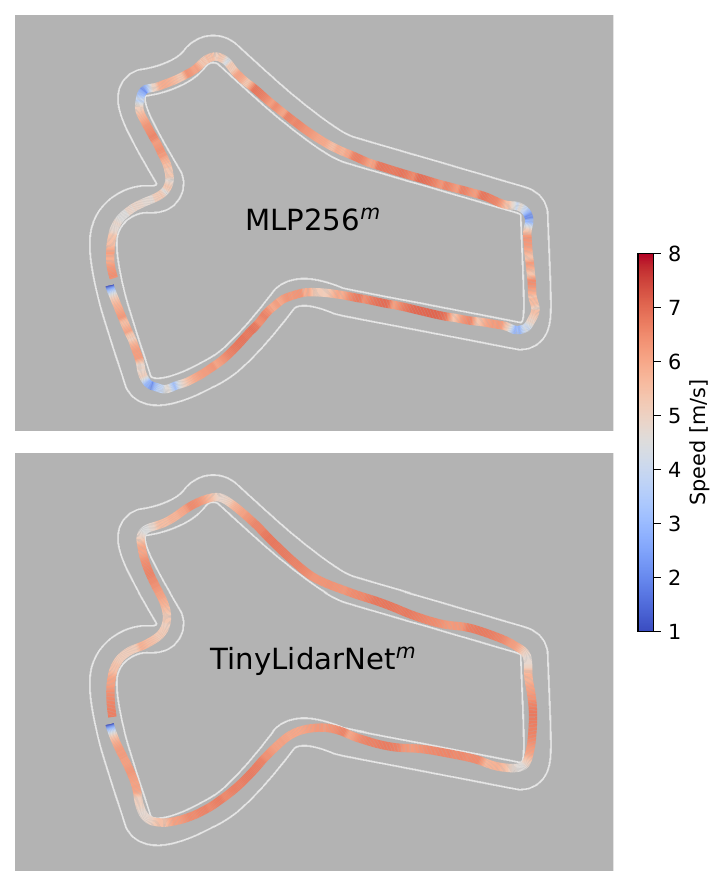}}
    \caption{A snippet of the trajectory of different models with speed profile on GYM track.}
    \label{fig:Trajectory}
\end{figure}


The Figure \ref{fig:Compare_diff_tech} and Figure \ref{fig:Trajectory} show the driving speed and the trajectory, respectively, of the TinyLidarNet$^M$ and MLP256$^M$ models in one of the successfully completed laps out of 10 trials on the GYM track. Note that MLP256$^M$'s speed fluctuates significantly and its trajectory is wobbly. In contrast, TinyLidarNet$^M$ maintains a more consistent speed and stable trajectory. Note that in terms of average progress, all TinyLidarNet models consistently outperform MLP256 models in all evaluated tracks. Regarding average lap time, TinyLidarNet models are usually faster than or similar to MLP256 models. 



In summary, we show that TinyLidarNet models, trained on real-world data, are well generalized to navigate unseen race tracks, while the commonly used MLP models struggle to generalize. This is because the CNN layers of TinyLidarNet can capture spatial features of LiDAR scans, enabling better generalization.  

\subsection{Inference Latency} \label{sec:eval-inference_latency}

In the context of racing, latency is of paramount importance due to the inherent stringent real-time requirements in such applications. In this subsection, we evaluate the inference latency of TinyLidarNet models on different computing platforms.

\begin{table}[h]
    \centering
    \begin{tabular}{lcccc}
    \toprule
         & \textbf{Xavier NX} & \textbf{ESP32-S3} & \textbf{RPi Pico} \\
    \midrule
        \textbf{CPU} & \makecell{NVIDIA Carmel\\6C@1.9 GHz} & \makecell{Xtensa LX7\\2C@240 MHz} & \makecell{ARM Cortex-M0+ \\2C@133 Mhz} \\
        \textbf{Memory} & 8GB LPDDR4x & 8MB PSRAM & 264KB SRAM \\
        \textbf{Storage} & 16GB eMMC & 8MB Flash & 2MB Flash\\
    \bottomrule
    \end{tabular}
    \caption{Computing platforms}
    \label{tab:board_comparision}
\end{table}

Table~\ref{tab:board_comparision} shows the three computing platforms we use for evaluation. Note that Jetson Xavier NX is the main onboard computer platform of our F1TENTH platform, which features powerful six ARM CPU cores running at 1.9 GHz. We also use XIAO ESP32-S3~\cite{esp32s3} and Raspberry Pi Pico~\cite{pico} MCUs to understand the potential of using these inexpensive low-end computing platforms for the real-time processing of TinyLidarNet. To this end, we port all three versions of TinyLidarNet on the MCUs and measure the inference latency of the models. 


\begin{table}[htbp]
    \centering
    \begin{tabular}{lccc}
    \toprule
    \textbf{Model} & \textbf{Xavier NX} & \textbf{ESP32-S3} & \textbf{RPi Pico} \\
    \midrule
    TinyLidarNet$^L$ (fp32) & $<$1 & 838 & 2642 \\
    TinyLidarNet$^L$ (int8) & $<$1 & 16 & 196 \\
    TinyLidarNet$^M$ (int8) & $<$1 & 8 & 91 \\
    TinyLidarNet$^S$ (int8) & $<$1 & 4 & 36 \\
    \bottomrule
    \end{tabular}
    \caption{Inference latency (ms) comparison}
    \label{tab:performance_transposed}
\end{table}

Table~\ref{tab:performance_transposed} shows the results. Note that (fp32) refers to the baseline 32-bit floating point number-based models whereas (int8) refers to 8-bit integer quantized models to reduce computational overhead and storage space demand to store the weights of the model.

First, note that all TinyLidarNet models are small enough to fit in both MCUs. 
Even in the smaller Raspberry Pi Pico MCU with only 264KB SRAM and 2MB Flash, all TinyLidarNet models can fit 
without any issue.

Second, the inference latency of TinyLidarNet$^L$ (fp32), however, is more than 2 seconds on the Pico and more than 800 ms on the ESP32-S3 MCU, which is more powerful than Pico but still significantly limited compared to Xavier NX. 

Third, using 8-bit quantized models enables TinyLidarNet to be executed in real-time as the TinyLidarNet$^L$ (int8) model takes only 16 ms to perform an inference on the ESP32-S3. This means that the model can run at $>$50Hz. The inference latency of the smallest TinyLidarNet$^S$ (int8) model is only 36 ms ($>$20Hz) on the Pico, which is sufficient for real-time control for most robotics applications.  

In summary, with a standard quantization-based optimization, we show that TinyLidarNet is lightweight enough to run on low-end MCUs in real-time.

\subsection{Performance on Unseen Real Tracks} \label{sec:MLP}

In this subsection, we evaluate the performance of TinyLidarNet on an unseen real track to validate the findings on the simulated tracks in Section~\ref{sec:eval-generalizability}. 

As in Section~\ref{sec:eval-generalizability}, the evaluated models are all trained using the 
dataset we collected during the F1TENTH competition, described in Section~\ref{sec:TinyLidarNet_Training}. 
The trained models are then tested on a different track installed in our lab. 
Figure~\ref{fig:12th_f1tenth_track} shows the track. 

\begin{figure}[htp]
  \centering
 \includegraphics[width=\linewidth]{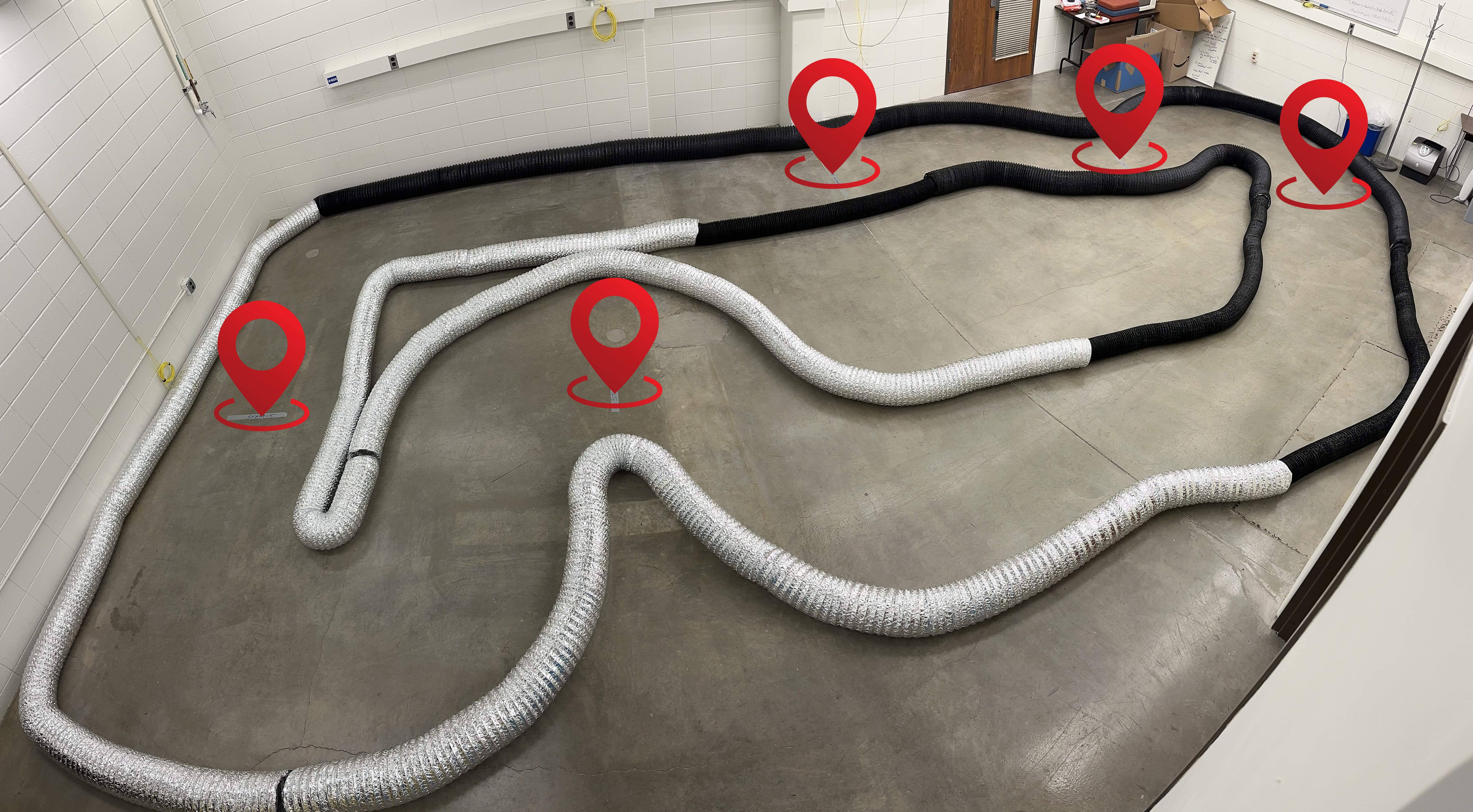}
  \caption{Real-world test track}
  \label{fig:testing_track}
\end{figure}

We employed a similar performance metric to what was discussed in Section \ref{sec:eval-generalizability}, conducting these experiments over 5 trials with the ego vehicle placed on the track at 5 randomly selected starting positions, indicated by the red markers visible in Figure \ref{fig:testing_track}. To ensure a fair comparison, we set the minimum speed to -0.5 m/s, considering the vehicle's inertia, and capped the maximum speed limit at 5 m/s, aligning with the linearly mapped data range from 0 m/s to 5 m/s used during training for all models.



\begin{table}[htp]
\centering
{%
\begin{tabular}{lcccc}
\toprule
\textbf{Model} & \multicolumn{1}{c}{\textbf{\makecell{Average \\ Lap Time (s)}}} & \multicolumn{1}{c}{\textbf{\makecell{Success \\ Rate (\%)}}} \\ 

\midrule

TinyLidarNet$^{L}$     & 20.5 & 100 \\ 

TinyLidarNet$^{M}$     & 19.9 & 100  \\ 

TinyLidarNet$^{S}$     & 19.5  & 80 \\ 

\hline


MLP256$^{L}$~\cite{sun2023benchmark}  & N/A & 0 \\ 
MLP256$^{M}$  & N/A & 0 \\ 
MLP256$^{S}$ & N/A & 0 \\ 





\bottomrule
\end{tabular}%
}
\caption{Average lap time and success rate of the models on the real track. The average lap time was calculated from the successful lap completion of 5 trials by placing the ego vehicle in random positions on the map.}
\label{tab:real_world_performance}
\end{table}

Table~\ref{tab:real_world_performance} shows the results. 
Across five trials, all MLP256 variants consistently crashes, highlighting the struggle to generalize on a new track that was not seen during training. 
All three MLP256 models were unable to complete a single lap and struggled especially when they needed to take hard U-shaped turns at the bottom left and top right corners of the track shown in Figure \ref{fig:testing_track}.

In contrast, all three versions of TinyLidarNet were able to complete the lap.  
TinyLidarNet$^{S}$ crashed only once (out of five trials), possibly because it is the smallest of all and loses 3/4 of the LiDAR range information due to downsampling (skipping three out of every four range values). Nevertheless, all TinyLidarNet models clearly outperform the MLP model. 

As discussed earlier, TinyLidarNet's superior performance can be attributed from its use of 1D CNN layers, which can capture spatial features of 2D LiDAR scans better. As a result, TinyLidarNet models generalize well on unseen tracks as well as unseen moving vehicles, all of which have distinct spatial features that can be learned. While it is well-known that CNN is good at processing vision data, our results strongly suggest that it is also good at processing 2D LiDAR data, which is consistent with recent findings in ~\cite{bosello2022train}.

%% file: sections/conclusion.tex
\section{Conclusion}\label{sec:conclusion}

In this study, we introduced TinyLidarNet, a lightweight 2D LiDAR-based end-to-end deep learning model designed for F1TENTH racing. TinyLidarNet takes an array of LiDAR scans as direct input and predicts steering and speed through a 1D convolutional neural network. 

We conducted a comprehensive evaluation on the generalizability of the model on unseen tracks both in simulation and in the real-world. We showed that TinyLidarNet's 1D CNN-based architecture is superior to the commonly used MLP architecture in processing 2D LiDAR scan data. 


We also evaluated the effects of quantization on TinyLidarNet. Deploying it on two low-cost MCUs, we found its inference to be sufficiently efficient for autonomous vehicle control, demonstrating its lightweight nature and low latency.


Future research avenues include exploring the use of DAgger~\cite{ross2011reduction,kelly2019hcdagger} to improve data collection, and using TinyLidarNet as a foundation for bootstrapping with Deep Reinforcement Learning techniques,  further enhancing its capabilities and adaptability in challenging racing scenarios.

%% file: sections/acknowledgement.tex
\section*{Acknowledgments}\label{sec:acknowledge}

This research is supported in part by the NSF grant CPS-2038923.